# AI-driven Structure Detection and Information Extraction from Historical Cadastral Maps (Early 19th Century *Franciscean Cadastre* in the Province of Styria) and Current High-resolution Satellite and Aerial Imagery for Remote Sensing

Wolfgang Göderle[1], Christian Macher[2], Katrin Mauthner[2], Oliver Pimas[2], Fabian Rampetsreiter[1]

Cadastral maps, which were produced from the early 19th century onwards for large parts of Europe under meticulous quality standards, represent a unique and complex representation of knowledge (W. T. Göderle 2017; W. Göderle 2023; Wheeler 2023; Femenia-Ribera, Mora-Navarro, and Pérez 2022). In the case of Habsburg Central Europe (where cadastral mapping took place between 1816 and 1861), they contain detailed information on every single house at a very large scale (between 1:720 and 1:5760, standard 1:2880), on land use, on contemporary roads and ways, the location and boundaries of land, and further data that could be extracted (N.N. 1824; Bundesamt für Eich- und Vermessungswesen 2017: 44). Beyond their immense value as a historical source, they can also be related to contemporary spatial representations for important questions, be it environmental and transport history, building history, economic history and social history, further questions of land use and the transformation of landscapes in the Anthropocene, and for obtaining information on resource extraction at the time of map creation (Ståhl and Weimann 2022).

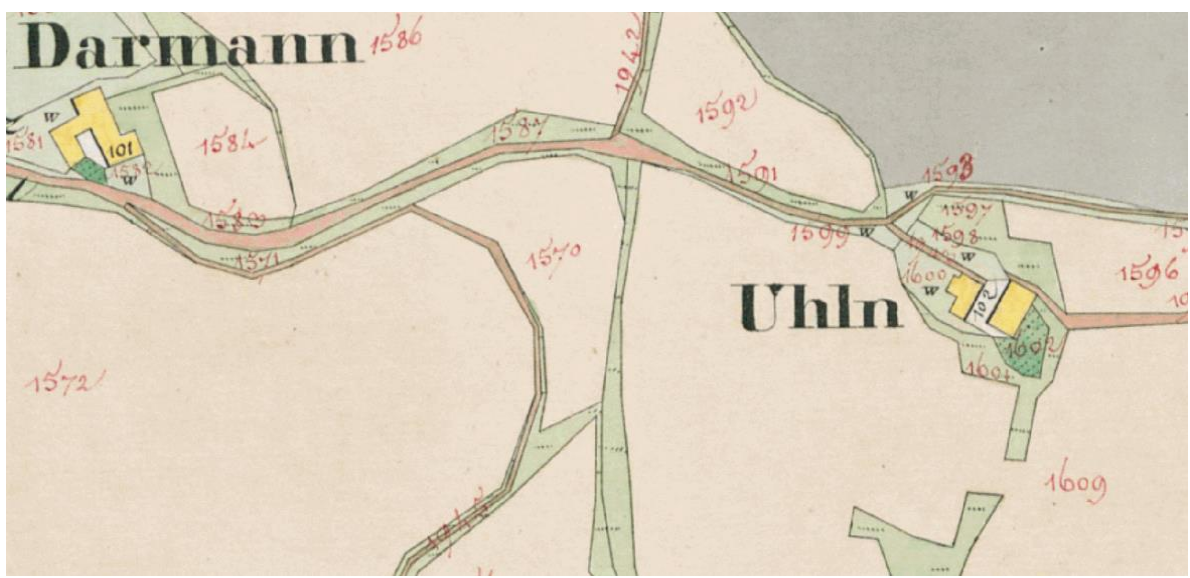

[1] University of Graz, History Department
[2] Know Center, Graz

*Figure 1: Extract from cadastre from 1828, wooden houses are visible in yellow, along with paths and agricultural land.*

Other aspects, however, make the cadastre a difficult source to access and process. Its multimodal structure, and its primary focus on image and geodata, pose a challenge for the digital humanities, whose central innovations lie in the textual domain(Champion 2017; Van Noord 2022).

In many cases, digitisation rendered cadastres accessible to a wider audience (Bundesamt für Eich- und Vermessungswesen 2017; Femenia-Ribera, Mora-Navarro, and Pérez 2022; Hagemans et al. 2022). Although archaeology has been using cadastral maps for a long time, historians and researchers from other disciplines have only been able to access the extremely extensive volumes stored in a few archives in exceptional cases. Due to the extent of this knowledge representation, its content could only be made accessible in part in pre-digital times. Digitisation has also not yet led to a comprehensive scaling of the evaluation of the material contained in the cadastre, in the case of Habsburg Central Europe: The 300,000 square kilometres of Central Europe surveyed between 1816 and 1861 in the course of the cadastral survey yielded almost 165,000 individual map sheets recording every place, every house, every road, every plot of land and the respective land use(Bundesamt für Eich- und Vermessungswesen 2017). A detailed evaluation beyond the local level is tedious and requires a great deal of manpower.

The FWF project RePaSE – Reading the Past from the Surface of the Earth – 2022 set out to test the AI-assisted extraction of building outlines and locations from geo-referenced map sheets of the *Franziszeische Kataster* (Franciscean Cadastre). An AI model was trained for this purpose. In a second step, another model was trained to extract buildings multimodally from high-resolution aerial and satellite images. In a third and final step, the two layers, where one contained information on the position of historical buildings, and the other information on present buildings, were superimposed and used to locate sites of possible archaeological relevance, in order to significantly facilitate archeological remote sensing.

The hypothesis behind RePaSE is that by matching information from two sources – once historical, from cadastral maps, and then currently from satellite and aerial photography – one could theoretically identify for entire regions those places where buildings stood two hundred years ago that have since disappeared. This could significantly support the identification of potentially archeologically important sites at scale.

## State of Research

Access to historical cadastral maps became publicly accessible only relatively late in Central Europe (Pivac et al. 2021). Presumably since 2016, map data is widely available via the www.mapire.eu project, however, the repository features serious shortcomings with regard to available metadata, further, data cannot be downloaded used in research contexts free of charge. Public authorities have rendered different parts of the *Franziszeische Kataster* online meanwhile, usually on provincial, sometimes on national level, and the data is usually provided free of charge in different qualities for research purposes (Pivac et al. 2021).

Lately, the exploration of historical map data in general and cadastral map data in particular at scale with the assistance of appropriate tools has accelerated considerably, due to recent advances in machine learning. Particularly the strong performance of deep learning-based technologies has led to significant gains in different specific fields (L.-C. Chen et al. 2016). Feature extraction, which is frequently focusing on either streets (Ekim, Sertel, and Kabadayı 2021; Can, Gerrits, and Kabadayi 2021; Jiao, Heitzler, and Hurni 2021; Uhl et al. 2022; Jiao, Heitzler, and Hurni 2022) or buildings (Heitzler and Hurni 2020; Uhl et al. 2020) for instance, yet also in a more general sense (Rémi Petitpierre, Kaplan, and Di Lenardo 2021; Garcia-Molsosa et al. 2021) have opened up a wide array of new possibilities for researchers from several relevant areas, history, archeology, and historical geography among them.

An important body of research has emerged in the field of feature extraction specifically from cadastral maps, particularly with regard to the Venetian cadastre(Oliveira et al. 2019), yet also the 1900 Atlas of Paris(Remi Petitpierre and Guhennec 2023). As Petitpierre and Guhennec point out, consistent annotation is the main challenge with regard to automatized vectorization.

Even though maps and further historical maps attract significant scholarly and scientific attention lately(Jiao, Heitzler, and Hurni 2022), the bulk of research is going into the analysis of aerial and satellite imagery(J. Li et al. 2022; Ji, Wei, and Lu 2019; M. Li et al. 2023; Fiorucci et al. 2020; 2022; Bickler 2021; Ren et al. 2015; Ding and Zhang 2021; Luo, Wu, and Wang 2022; Lee, Wang, and Lee 2023; K. Chen et al.

2021). We observe a wide array of different objectives and interests here, and stakeholders ranging from municipal and urban administrations to archeological remote sensing missions. An important body of work is focusing on LIDAR data (Fiorucci et al. 2022), further, the transformation of urban landscapes and the changes in the extent of urban areas triggers relevant research (Uhl et al. 2021), as well as realtime-surveillance tasks that frequently involve UAVs (Ren et al. 2015; Ding and Zhang 2021; Luo, Wu, and Wang 2022; Lee, Wang, and Lee 2023).

## Task

The larger objective of RePaSE was twofold: RePaSE should provide a proof-of-concept that AI-assisted extraction of large amounts of data – in our use case: the location and the form of buildings – from the cadastre is already possible with existing resources and, in the best case, identify suitable models to that purpose. The second step was to make RePaSE fit for future multimodal applications, and to identify and test models that could be used to extract the location and the form of buildings also from current aerial and satellite imagery that is already available (Smits and Wevers 2023).

We therefore set up the following research strategy: The research task was divided into two sub-tasks, task 1 being made up of the detection of objects, buildings, in scans of historical cadastral map data. Task 2 consisted of the detection of objects, buildings in current satellite and aerial imagery. The output layers of task 1 and task 2 were finally superimposed, as we were specifically interested in places that had buildings in the cadastre but no longer in the present. Overlaying automatically extracted location data from the cadastre and those from the present resulted in the relevant locations for us to be marked in a layer. The latter was used for analysis with different spatial representations, for instance map data, yet particularly satellite and aerial imagery.

## Approach

As it turned out that the size of the annotated map patches had a significant impact on the performance of our models, we resized all map patches to the uniform size 3747x2235, which was a dimension, which turned out to be working quite well. However, we employed three different zoom levels – close, medium and far.

We used the software tool cvat to annotate the map patches ("www.cvat.ai," n.d.).

The first task to be addressed was an image segmentation problem, in order to be able to successfully extract houses from cadastral map material. In our first attempt, we followed a low-key approach and tried clustering algorithms to tackle this challenge, which turned out unsuccessful. In line with the fail-fast approach of our proof-of-concept study, we subsequently turned to a completely different method. When it comes to deep neural networks in this area of application, convolutional neural networks (CNN) have outperformed other architectures in the last years(Rawat and Wang 2017; O'Shea and Nash 2015). As building extraction can be considered a computer vision problem, fully convolutional networks are considered the most widespread state-of-the-art solution to this kind of problem (K. Chen et al. 2021). We therefore chose the Google DeeplabV3 model (L.-C. Chen et al. 2017), which is considered to be very efficient in dealing with the multi-scale problem featured in this context, due to its atrous convolution layer (L.-C. Chen et al. 2016). We finetuned DeepLabV3 with 50 annotated example images that we split into 6 squares each to give 300 images in total. The images were split into training, testing and validation set according to the following scheme:

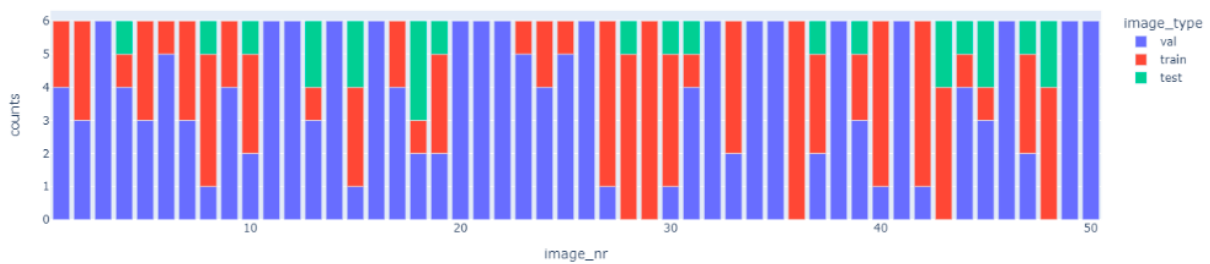

*Figure 2: Finetuning the DeeplabV3 model, color code: blue – validation data; red – training data; green – test data.*

Squares that did not contain any building were automatically assigned to the validation set (that is why there are so many in that set compared to test and train set), in order to enable the model to learn that not every map patch necessarily contains buildings. The three datasets train, test and val only contained unique images, each image featured only once over all three datasets.

Finetuning was based on a repository that had been developed for transfer learning for semantic segmentation purposes (Singh and Popescu 2021). Building on a DeeplabV3 with resnet 101, which formed the basis for our cadastre deeplabv3, we finetuned over 20 epochs (using a server and regular workstations), until we yielded satisfactory results("Https://Pytorch.Org/Vision/Stable/Models/Generated/Torchvision.Models.Seg

mentation.Deeplabv3_resnet101.Html#torchvision.Models.Segmentation.Deeplabv3_resnet101" 2017).

The second task to be addressed was another image segmentation problem, the extraction of buildings from high-resolution aerial and satellite data. As the quality of satellite imagery freely available has once more increased in the two years between the setup of the project idea and the project execution, we could use satellite imagery in better quality than originally expected, yet we could also add very-high-resolution aerial photography and LiDAR data to the research data under scrutiny. Unlike the 19th century cadastral map data that we were dealing with in task 1, we were addressing an entirely different problem here, as buildings in ortho- and satellite imagery can take on a wide range of very different appearances. Due to the significantly different challenge, we chose an adapted approach. Vision transformers – thus transformer architectures adapted for the use in computer vision tasks – are considered to feature relatively economic computational overhead and memory consumption, while they can achieve state-of-the-art accuracy (K. Chen et al. 2021). We therefore trained a sparse token transformer (STTnet) from scratch, using a GPU. STTnet is a state-of-the-art vision transformer with very promising results in comparable tasks(K. Chen et al. 2021). Building on Resnet50 as backbone, we selected 2.657 labelled images from the AIRS dataset. This was so that we could avoid labelling houses in the satellite images ourselves. The effort for this would have been considerable, because for training from scratch we need substantially more training examples than for finetuning. All images resembled the Alpine topography and the landscape characteristics of our target region Styria. We finally used imagery that was taken in Austin, Chicago, Kitsap (Washington State), Tyrol and Vienna. The training took place over 98 epochs, once the loss rate and the F1 score did not feature significant changes anymore, the training was considered completed. The model was then tested with 224 labelled images.

Subsequently, the extracted layers were superimposed and negative profiles were created. The negative profiles contain archaeologically potentially interesting locations, which once saw buildings, yet that are not overbuilt at present.

## Results

The results for task 1 – the extraction of buildings from cadastral maps via DeeplabV3 turned out very well.

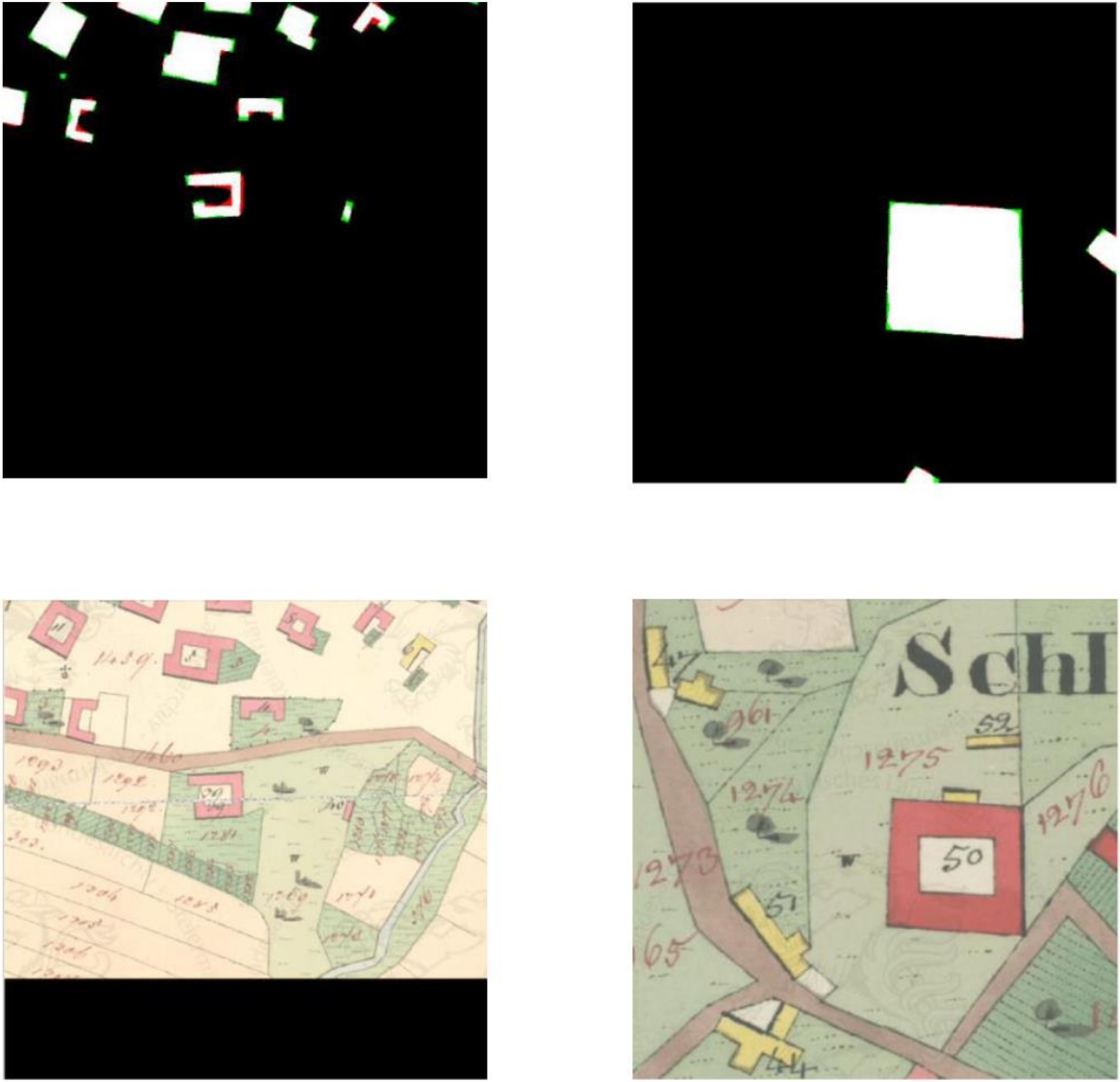

*Figure 3: Building extraction from cadastral maps with finetuned DeeplabV3*

As can be seen from the images, the extraction of red houses (stone houses), for which the model was finetuned worked impeccably, which is reflected in the metrics we obtained. We calculated the parameter intersection over union (IoU), where the labelled mask – thus the image that we prepared for training by labelling – is compared with the predicted mask, thus the result the model actually delivers once the training is completed. The value is calculated per class, whereas we defined two classes here, house (class 1) and background (class 2). We calculated two values:

Micro average = $\frac{Intersection\ House + Intersection\ Background}{Union\ House + Union\ Background}$

Macro average = $\frac{IoU\ House + IoU\ Background}{2}$

We yielded a mean micro average IoU of 0.990389 and a mean macro average IoU of 0.89516, which represent outstanding values. The accuracy of the model was thus sufficiently high. Among the problems we encountered were some streets that were mistaken for houses, and minor differences in the color schemes (shades) in the hand-drawn maps. These were, however, minor issues, which we believe that we could deal with quite simply by labelling more training examples and retraining the model. Further finetuning could teach the model to recognize further structures, particularly yellow houses (wooden buildings) and streets, which will be our priority in a follow-up project.

The situation with task 2 was slightly different. The achieved IoUs of 0.537755 (macro) and 0.795838 (micro) proved to be satisfactory in this context from a pragmatical perspective, as we were well able to work with the obtained results. Although these are significantly below the values achieved for task 1, we consider this task successfully solved. The building outlines contained in the negative layer must be kept more generous anyway, as it is evident, especially in the terrain model, that the terrain is affected by the development beyond the actual building boundaries, through the levelling of the built-over area. The red markings in the upper layer mark false positives (a house was detected, where there was no house in reality), the green markings false negatives (the model failed to detect the house). White indicates a correct identification. The model sometimes displays difficulties correctly identifying houses at the margins of the detection area. Furthermore, it was difficult for the model to achieve good results outside a certain (optimal) zoom range. This problem could be circumvented by properly addressing the issue in the data preprocessing. As a rule, the model tended to minimally enlarge the recognised buildings, which, however, was quite convenient for our work, as already noted. From an academic perspective, we see some possibilities to optimize the model. We are certain that additional training, i.e. adding more epochs, would yield a substantial improvement of the performance.

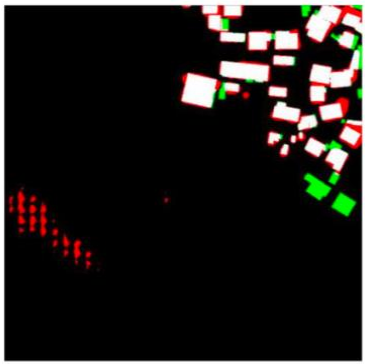
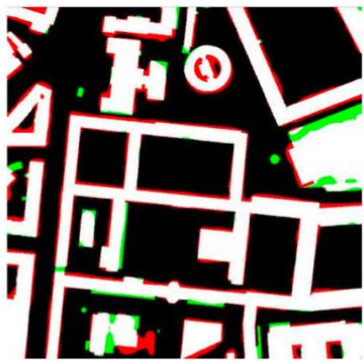
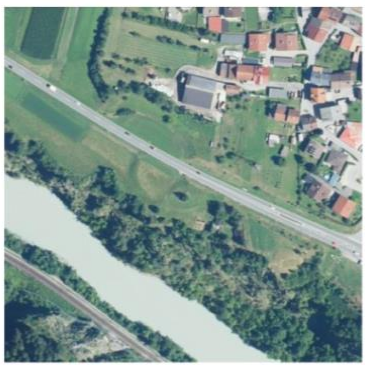
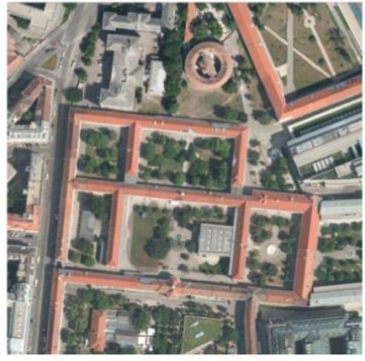

*Figure 4: STT results Tyrolia (left)*     *Figure 5: STT results Vienna (right)*

## Next steps

RePaSE provided proof-of-concept with regard to AI driven data extraction from historical map material and extremely valuable experience in terms of a way forward. The most important finding is that structures can be extracted from historical cadastral map material well and economically in very high quality. This opens up a wide range of new possibilities for us. Therefore, this is a field, in which we plan on engaging further, building on the experience we gathered in RePaSE and further in cooperation with several other groups working in this field of research. We expect that particularly the highly relevant findings of colleagues concerning the synthetic production of suitable training data (Jiao, Heitzler, and Hurni 2022) will allow for a significant breakthrough in extracting relevant data from historical maps at scale.

In view of the rapid progress in the extraction of graphic structures from high-resolution satellite and aerial photographs, it seems sensible to orientate on the current state of research instead of investing own resources in improving the technology. Nevertheless, we consider it important to keep an eye on this field, as digital humanities and humanities data science struggle to access state-of-the-art technologies, models and datasets to process information explainable. We consider the solution we identified in the context of RePaSE – STTnet – viable and expandable, though it will require a substantial effort to toughen up our outputs to a level, at which the quality of results will be robust enough to qualify for positive profiles. The STTnet we trained is perfectly working in a project context such as RePaSE, it will however require some adjustments to make it work in a different context where a higher IoU metric might be required.

We have identified structures that can be detected very well when using computer vision, and which bear important time signatures, thus, which actually promise to render the past that is hidden in the surface of the Earth visible to the eyes of researchers. Further, as we made promising and encouraging experiences with other object detectors on related tasks recently, we believe to have identified promising paths of development in this field.

A follow-up to RePaSE is currently in development, it will fully integrate the two models trained and finetuned within RePaSE – DeeplabV3 and STTnet – and will set out to identify and train two further models with a focus on historical road infrastructure.

## Conclusion

RePaSE provides a wide variety of stakeholders, from municipal authorities over political decision makers at the local and regional level to historians and archeologists with a powerful tool to detect sites of potential archeological relevance, as is shown on the following two figures:

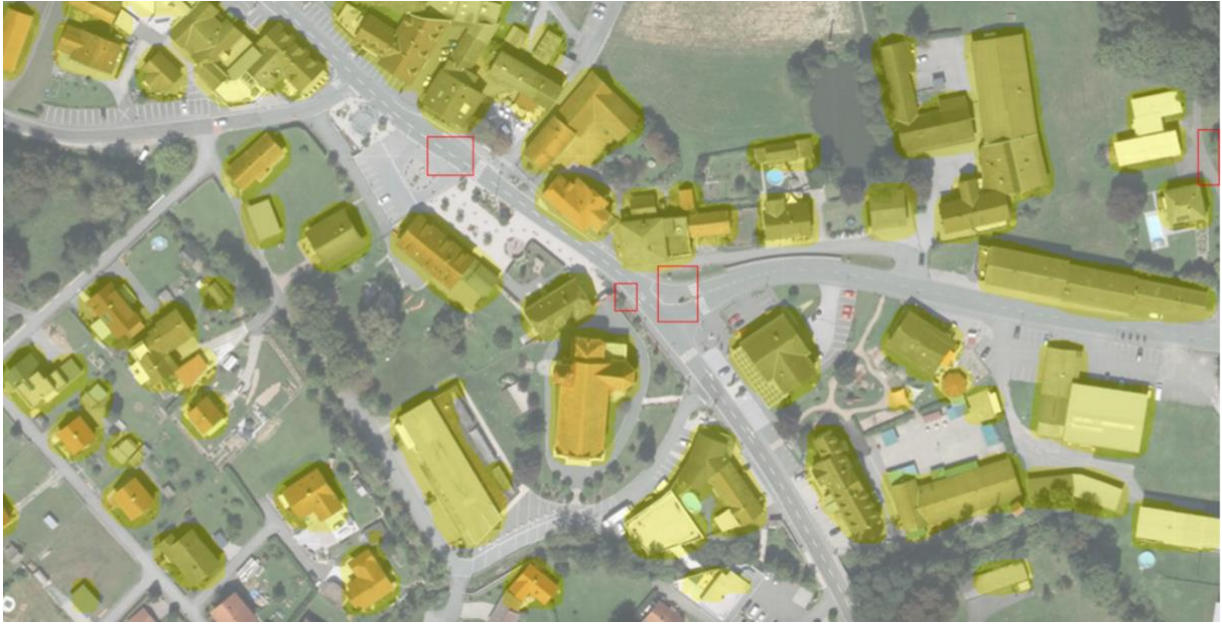

Figure 6: Present-day high-resolution aerial image of a part of the rural municipality Heiligenkreuz. Bounding boxes mark automatically detected locations of buildings, which featured in the 1820s cadastre. The yellow shade marks the identified present-day buildings.

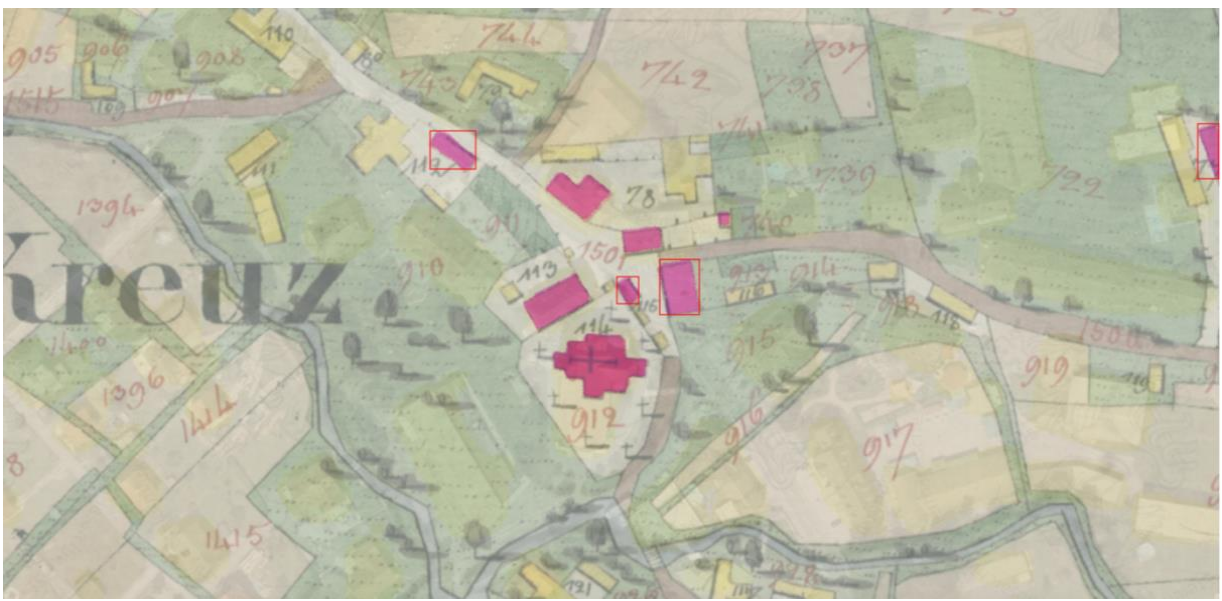

Figure 7: The respective map patch in the cadastre, the buildings that have since disappeared feature in the bounding boxes.

The RePaSE-team is currently working to set up a server to provide full access to the above-mentioned group to the entire dataset processed in the project runtime. RePaSE is currently working for the entire region of Styria, we expect that it could easily be expanded for the adjoining regions of Carinthia, Lower Austria, Salzburg and Slovenia. As Croatia and Burgenland were part of the Transleithanian cadastre, which features slightly different optical features, we expect to meet some difficulties there, yet the system would probably be working with another round of finetuning.

RePaSE clearly showed the enormous potential of machine learning, particularly with regard to computer vision, image segmentation and object detection to scale up analysis of historical map material and to combine these insights with (AI harvested) data gained from current high-resolution imagery of the surface of the Earth. Models of the latest generation feature outstanding usability and require manageable effort to successfully extract data from large-scale knowledge representations. It turned out in the context of this project, though, that domain knowledge plays an increasingly important role.

Whereas image segmentation must be regarded a standard technology with regard to this type of sources now, due to its good availability, usability and accessibility, possibly the most relevant result of RePaSE with regard to the question, what digital humanists' tools will be for working with these data in the near future, is: Based on our results, what we learnt is essentially that object detection bears enormous potential for many crucial tasks, particularly when it comes to the analysis of historical map data at scale. The object detectors we have been working with so far display a wide range of possible applications, which reach far beyond the purpose they were developed for. The fact that many object identification models combine excellent object recognition with relatively manageable hardware requirements makes them eminently suitable tools for use in the DH context.

It is already possible to train a Faster R CNN model for a specific use case under economically reasonable conditions (Ren et al., 2015), with which the results of massively pre-trained, fine-tuned large models can be outperformed. The transfer of his method is realistic in its basic features for the problem described in RePaSE and will be tested in the next step.

**Notes**

1. RePaSE was funded by the Austrian Science Fund (FWF) under the reference TAI 591.
2. Further information on RePaSE and access to the demonstrator already available: https://repase.uni-graz.at. The demonstrator can be accessed via https://repase.know-center.at.
3. We were granted access to the highest-resolution research data hosted by https://gis.stmk.gv.at/wgportal/atlasmobile. Similar research data are available for all nine provinces of Austria., for example for Carinthia under https://gis.ktn.gv.at/webgisviewer/atlas-mobile/. Unfortunately, comparable access is not easy in all successor states of the Habsburg Monarchy. The arcanum.com service offers good usability and orientation possibilities, but no metadata is offered free of charge that would enable scholarly work, and the resolution is limited.
4. Further relevant repositories and projects focusing on the cadastre can be found under www.franziszeischerkataster.at. Further, the project HiLaK – Historische Landnutzung als Grundlage für Klimaschutzmaßnahmen heute, directed by Kurt Scharr, and its follow-up project HiLuC display state-of-the-art approaches to cadastre-based multi-disciplinary research in the field with regard to Austria.
5. We will provide all datasets that we used and produced to the scientific community as soon as the project has been concluded administratively, which will most probably be the case in Q1/2024.

ABSTRACT

Cadastres from the 19th century are a complex as well as rich source for historians and archaeologists, whose use presents them with great challenges. For archaeological and historical remote sensing, we have trained several Deep Learning models, CNNs as well as Vision Transformers, to extract large-scale data from this knowledge representation. We present the principle results of our work here and we present a the demonstrator of our browser-based tool that allows researchers and public stakeholders to quickly identify spots that featured buildings in the 19[th] century Franciscean Cadastre. The tool not only supports scholars and fellow researchers in building a better understanding of the settlement history of the region of Styria, it also helps public administration and fellow citizens to swiftly identify areas of heightened sensibility with regard to the cultural heritage of the region.